\documentclass{article}

\usepackage{microtype}
\usepackage{graphicx}
\usepackage{subcaption}
\usepackage{booktabs} %

\usepackage[ruled,algo2e]{algorithm2e}
\usepackage{float}
\usepackage{amssymb}
\usepackage{amsthm}
\usepackage{scalerel}
\usepackage{stackengine}
\usepackage[nointegrals]{wasysym}
\usepackage{tablefootnote}
\usepackage{wrapfig}
\usepackage{xfrac}
\usepackage[export]{adjustbox}
\usepackage[dvipsnames]{xcolor}

\usepackage{amsthm,amsmath,amssymb, amsfonts}

\usepackage{tabularx}
\usepackage{mathtools}
\usepackage{bbm}

\usepackage{anyfontsize}

\usepackage{hyperref}
\hypersetup{
    colorlinks=true,
    linkcolor=magenta,
    citecolor=blue,
    filecolor=magenta,      
    urlcolor=blue,
    pdfpagemode=FullScreen,
    }

\usepackage[nonatbib,preprint]{neurips_2021}
\usepackage[numbers]{natbib}

\usepackage{cleveref}

\newcommand\argmaxbigg[2]{\underset{#1}{\arg\max}\;\bigg(#2\bigg)}

\title{\mbox{General Intelligence Requires Rethinking Exploration}}

\author{%
Minqi Jiang\\
Meta AI \& UCL \\
\texttt{msj@meta.com} \\
\And
Tim Rockt\"{a}schel\\
UCL\\
\texttt{tim.rocktaschel@ucl.ac.uk}
\And 
Edward Grefenstette\\
Cohere \& UCL\\
\texttt{ed@cohere.com}
}

\begin{document}

\maketitle

\begin{abstract}    
We are at the cusp of a transition from ``learning from data" to ``learning what data to learn from" as a central focus of artificial intelligence (AI) research. While the first-order learning problem is not completely solved, large models under unified architectures, such as transformers, have shifted the learning bottleneck from how to effectively train our models to how to effectively acquire and use task-relevant data. This problem, which we frame as \emph{exploration}, is a universal aspect of learning in open-ended domains, such as the real world. Although the study of exploration in AI is largely limited to the field of reinforcement learning, we argue that exploration is essential to all learning systems, including supervised learning. We propose the problem of \emph{generalized exploration} to conceptually unify exploration-driven learning between supervised learning and reinforcement learning, allowing us to highlight key similarities across learning settings and open research challenges. Importantly, generalized exploration serves as a necessary objective for maintaining open-ended learning processes, which in continually learning to discover and solve new problems, provides a  promising path to more general intelligence.
\end{abstract}

\section{Introduction}

A hallmark of intelligence is a capacity to explore. From animals exploring their surroundings for food and shelter \citep{obrien1990search,gordon1999ants, manser2004spatial} to children seeking novelty in play \citep{piaget1952origins, piaget1954construction, spodek2014handbook, johnson2019play, yawkey2018child, andersen2022play}, exploration drives the acquisition of new information beneficial to the seeker. Exploration is even coded in the very fabric of life, in the form of genetic mutations that wander the space of phenotypes---some of which may improve the organism's chances to survive and reproduce. 
Through exploration, the explorer acquires not just information about a specific task or environment in isolation, but information in relation to previous experiences, informing the development of more generally-useful behaviors---which may include general strategies for searching for rewarding states in similar environments. Still, such information acquisition is not purely driven by the explorer. It is also determined by the learning opportunities provided by the environment: A static environment presenting a fixed challenge that the agent can already solve offers nothing more to learn. Similarly, a variable environment for which the agent already learned to explore optimally to find the solution offers no potential for learning new behaviors. Conversely, an environment presenting challenges far exceeding the capabilities of the agent is unlikely to provide sufficiently informative experiences for the agent to learn. 
Exploration should thus seek information most useful for the agent to become a more general and adept actor in future decision-making situations. In this way, exploration serves as the core mechanism for generating new experiences from which agents can learn to extrapolate or rapidly adapt to a changing world---to become a more generally-intelligent agent within the world.

The subject of generality in intelligence is a loaded concept to say the least. In broaching this subject, we must first clarify our treatment of general intelligence in the context of our discussion. As the set of tasks to which the notion of generality pertains is typically either undefined or infinite, definitions of general intelligence tend to be vague, incomputable, or unquantifiable in degree. Nevertheless, the quest for producing ever more general models has driven much progress in AI research \citep{schmidhuber2007godel, hutter2007universal, brown2020language, schrittwieser2020mastering, stooke2021open}. To avoid these definitional issues, we instead focus on a relative notion of general intelligence: Model $A$ is more general than model $B$ relative to a task set $\mathcal{T}$ if and only if $A$ performs above a threshold level (e.g. that of a minimally viable solution) in more tasks in $\mathcal{T}$ than $B$, while at least matching the performance of $B$ on all other tasks in $\mathcal{T}$ on which $B$ meets the threshold. Under this definition, general intelligence is not necessarily the end state of any system, but rather a property that can change over time, relative to other intelligent systems and the specific task domain. We can then refer to a system exhibiting continual improvements in relative general intelligence as an \emph{increasingly general intelligence} (IGI). Over time an IGI will overtake any other agent that is not an IGI in terms of relative general intelligence. For these reasons, our discussion conceives of ``general intelligence" as an IGI. In other words, we equate the goal of producing a general intelligence with producing an IGI. Moreover, for brevity, we exclude explicit reference to the associated task set on which IGI is defined. If not obvious given the context, the reader can assume we refer to the set of tasks which humans engage in, would engage in, or would benefit from. Further, we do not aim to address whether the notion of a general intelligence relative to the space of all possible tasks is well-defined, as it is an orthogonal concern to the argument we lay out in this paper.

\begin{figure}[t!]
    \centering
    \begin{subfigure}[b]{0.329\textwidth}
        \includegraphics[width=1\textwidth]{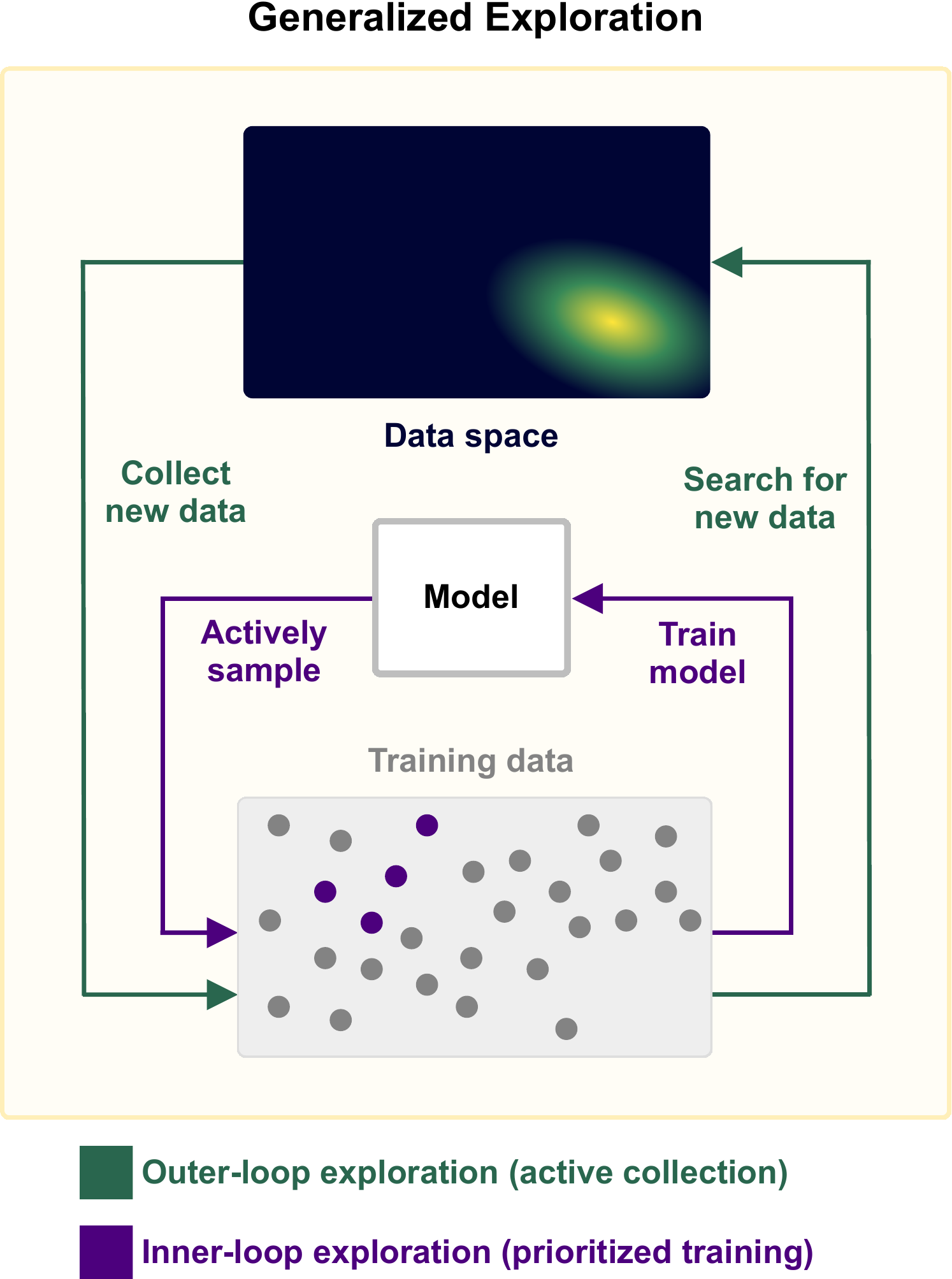}
    \end{subfigure}
    \begin{subfigure}[b]{0.329\textwidth}
        \includegraphics[width=1\textwidth]{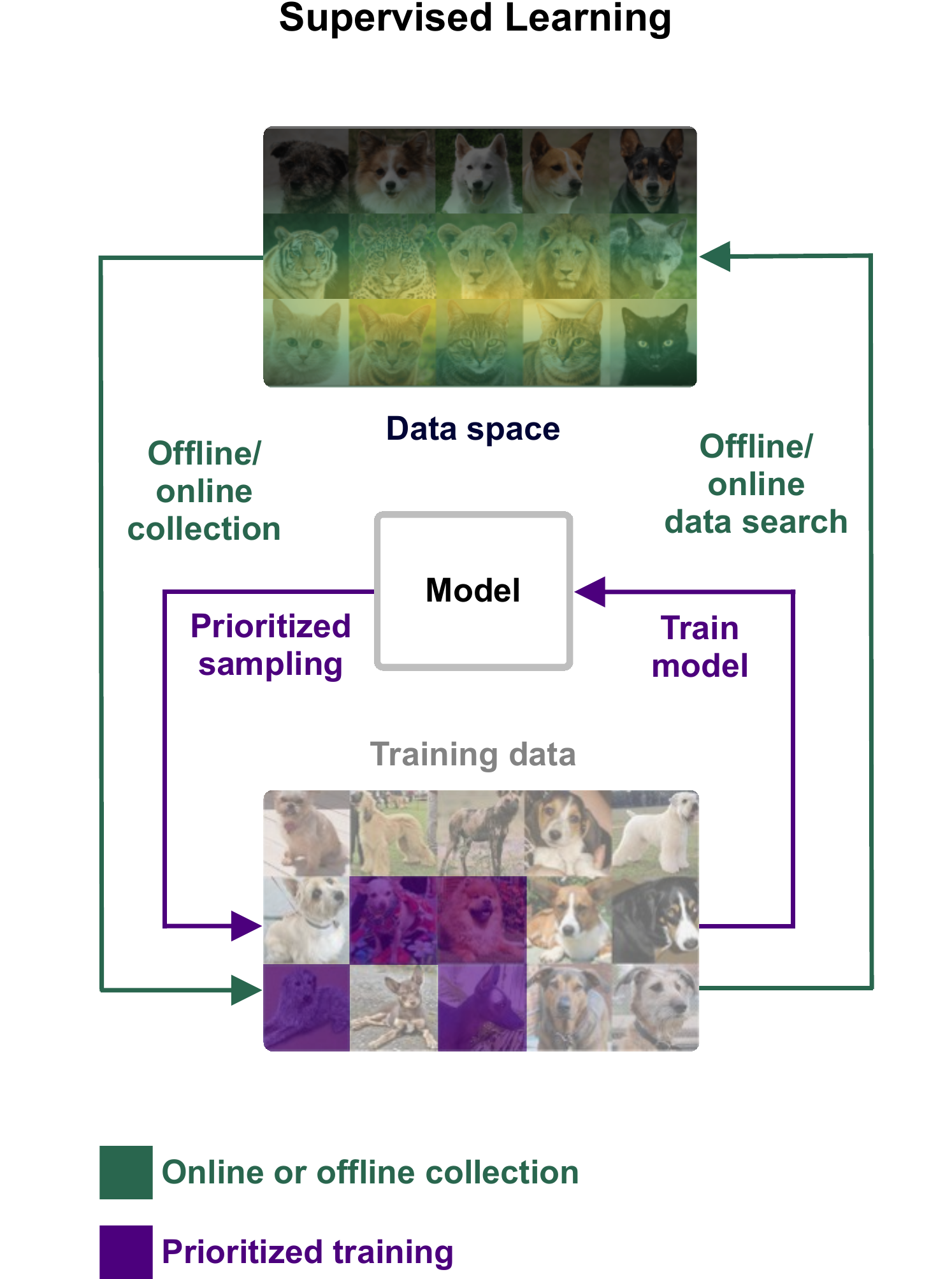}
    \end{subfigure}
    \begin{subfigure}[b]{0.329\textwidth}
        \includegraphics[width=1\textwidth]{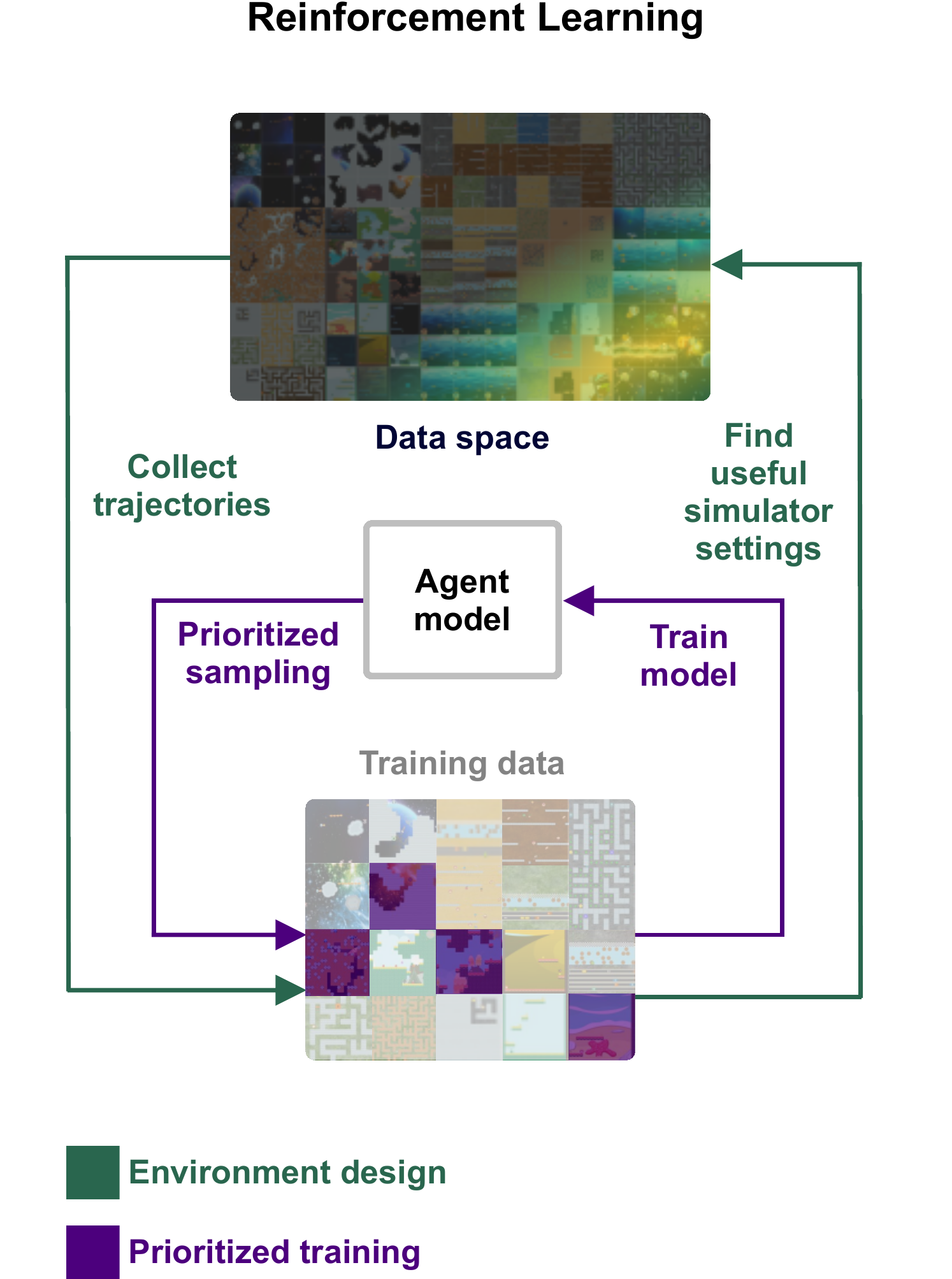}
    \end{subfigure}
    \caption{\small{A general framework for exploration: An outer loop performs active collection of new training data, and an inner loop conducts prioritized training on the current training data. In SL, the outer loop consists of either online or offline data collection. In RL, the outer loop searches for simulator settings that yield useful training data, and the inner loop can perform prioritized sampling, e.g. prioritized experience replay.
    }}
    \label{fig:unified_exploration_overview}
    \vspace{-8mm}
\end{figure}

A learning process leading to general intelligence should explore across two primary levels of abstraction: First, exploration within a specific environment facilitates the search for optimal solutions. Second, exploration across environments ensures continual learning progress and developing more general capabilities. If we assume the agent retains past solutions and continues exploring and discovering new challenges indefinitely, the agent can be expected to attain a broad spectrum of capabilities \citep{schmidhuber2007godel}. Importantly, the scope of capabilities should eventually surpass that resulting from any process with less effective exploration. Inversely, if the agent does not explore as thoroughly, we cannot expect it to acquire more capabilities than if it did. Therefore, increasingly general intelligence arises if and only if facilitated by open-ended exploration, which continually broadens the agent's capabilities. Such exploration ideally seeks new challenges at the \emph{boundary} of the agent's capabilities, focusing learning on tasks for which the current agent has the most potential to improve---analogous to Vygotsky's Zone of Proximal Development in developmental psychology \citep{vygotsky1978mind}.

In sharp contrast, the current research programs producing state-of-the-art machine learning (ML) systems, including large language-models (LLMs) \citep{brown2020language, hoffmann2022training, chowdhery2022palm} and generative models of images and video \citep{ramesh2022hierarchical, alayrac2022flamingo, ho2022video}, strive for generality without exploration. These large deep learning models are typically trained using variations of supervised learning (SL), including self-supervised learning \citep{oord2018representation}, and benchmarked on static datasets collected entirely offline. We argue that this paradigm, which we call \emph{offline supervised learning}, leads to models with unavoidable brittleness. Models trained in this way will exhibit blindspots reflective of the missing information in their training data and suffer from covariate shifts and concept drift due to the non-stationarity of the real world. 

Meanwhile, deep reinforcement learning (RL), which seeks to learn sequential decision-making strategies in simulated or real environments, directly considers the exploration-exploitation tradeoff, which treats exploration as a key objective, balanced alongside maximizing task-specific performance~\citep{sutton2018reinforcement}. RL acknowledges that task performance must often suffer in the short term in service to exploration—the only means to achieve even better  performance. Unlike SL models, which can take advantage of an effectively infinite trove of data amassed on the open web, the state-of-the-art RL methods remain limited to largely toy simulations or highly domain-specific applications. RL agents are largely limited to training in simulation for a number of reasons, although most typically this is due to the potential costs (e.g. financial, temporal, environmental) and dangers of real-world experience collection. As we argue in Section \ref{subsec:limits_rl}, current simulators predominantly mirror the limitations of a static, finite dataset in SL, and as such, RL agents share the same failure points as SL systems when deployed in the wild. Throughout this paper we will use the terms \emph{static environment} or \emph{static simulator} to refer to environments whose factors of variation form a constant set throughout the course of training. This definition encompasses both singleton environments, like each game in the popular Atari Learning Environment \citep{bellemare2013arcade}, which exist in the same configuration across every training episode, as well as environments like NetHack \citep{kuttler2020nethack}, which have a predefined set of configuration parameters whose values control specific attributes such as its appearance, topology, transition dynamics, and reward functions. Such attributes may take on different values in each training episode. While a static simulator with adjustable parameters can span a vast space of tasks, ultimately, it can only offer experiences within the limited domain that it was designed to simulate.

To introduce a case that will serve as a running example throughout this paper, consider a virtual assistant that responds to user queries through a chat-based interface and performs actions on behalf of the user by interfacing with other software, e.g.~answering questions by searching the web, finding and buying products, hailing a cab, or providing music and movie recommendations. Such an assistant could, in principle, learn to execute any task that can be performed digitally, which is becoming the majority of tasks humans in the information age are performing. Our assistant has the potential to become increasingly generally-intelligent, assuming it can sufficiently explore the space of tasks. Following our discussion, we see that current methods for training our assistant will not suffice: Training an increasingly generally-intelligent assistant requires continually exploring the vast and dynamically evolving space of tasks of interest to people, but current SL and RL methods are largely designed for training on a single, static dataset or simulator. An assistant trained with such approaches would quickly grow outdated with the onset of new world events and product offerings. The active subfield of continual learning \citep{hadsell2020embracing} aims to find ways to reliably retrain models on new data, but does not directly address the underlying question of how to determine and find the data that should be used for such updates in the first place.

We see that the problem afflicting both classes of learning algorithms reduces to one of insufficient exploration: SL, largely trapped in the offline regime, fails to perform any exploration, while RL, limited to exploring the interior of a static simulation, largely ignores the greater expanse of possibilities that the simulation cannot express. Thus, simply applying the current exploration paradigms developed in RL to SL is not enough to solve these problems. In order to properly address these limitations, we must first rethink exploration as it is currently framed in RL. We require a more general kind of exploration, which searches beyond the confines of a static data generator, such as a finite dataset or static simulator. This generalized form of exploration must deliberately and continually seek out promising data to expand the learning agent's repertoire of capabilities. Such expansion necessarily entails searching outside of what can be sampled from a static data generator. This open-ended exploration process, summarized in Figure~\ref{fig:unified_exploration_overview}, defines a new data-seeking outer-loop that continually expands the data generator used by the inner loop learning process, which itself may use more limited forms of exploration to optimally sample from this data generator. Within each inner loop, the data generator is static, but as a whole, this open-ended exploration process defines a dynamic, adaptive search process that generates the data necessary for training a likewise open-ended learner that, over time, may attain increasingly general capabilities.

Importantly, we argue open-ended exploration is necessary for collecting the most informative training data to feed our ever more powerful and data-hungry machine learning models. Previous studies suggest that, while model performance grows with the amount of training data, the quality of the data itself is of paramount importance \citep{zhu2016we, paul2021deep, hoffmann2022training, sorscher2022beyond}. As our understanding of optimization, modeling, and scaling laws improve in tandem with the continued expansion of computing power following extensions of Moore's Law \citep{moore1965moore, shalf2020future}, informative data is not only a limiting factor to more general models, but also an important counterweight to ensuring we do not overfit our models and algorithms to a limited subset of problems. Moreover, tailoring the sequence of training data presented to a model can greatly benefit sample efficiency and the quality of the learned representations. Continual exploration of informative data also serves as a critical means to ensure models remain well-aligned to an increasingly diverse world of dynamic human preferences and values. In this light, we argue a significant amount of research investment should shift toward the collection, design, and scheduling of training data itself---that is, toward developing principled methods for open-ended exploration. Combined with ever-improving model architectures and optimization techniques, such generalized exploration can lead to increasingly more general intelligent agents. 

Many prior works propose the potential of open-ended self-improving systems. Notably, both Schmidhuber \citep{schmidhuber2010formal, schmidhuber2007godel, schmidhuber2013powerplay} and Clune \citep{clune2019ai} have proposed novelty-seeking agents that generate their own challenges, while the work of Stanley and Lehman established the practical value of taming open-endedness \citep{lehman2011abandoning, stanley2015greatness, stanley2017open}. We do not claim to be the first to propose the concept. Rather, we seek to provide a detailed discussion of how open-ended learning can be integrated into modern ML systems and importantly, how this endeavor entails a meaningful generalization of prevailing notions of exploration. In the rest of this paper, we formalize the motivation and formulation of open-ended exploration. In Section~\ref{sec:limits_of_learning} we first build up a detailed discussion about the challenges that must be addressed by a successful framework of exploration for both SL and RL. An especially important challenge that motivates the need for continual exploration is the bootstrap problem in online learning, which we discuss in Secton~\ref{sec:bootstrap_problem}. We then propose a unified framework for exploration in Section~\ref{sec:exploring_beyond}, generalizing its application across problem domains and thereby providing a common framework for thinking about data collection in both SL and RL. Section~\ref{sec:unified_view} provides a discussion of how such open-ended learning may take place in bath practice and in research. As such open-ended learning provides a promising path to general intelligence, we relate these ideas to other proposed approaches achieving this grandiose goal in Section~\ref{sec:bottom_up_agi}. Finally, in Section~\ref{sec:open_problems}, we discuss what we consider the key open problems to realizing this path to general intelligence.

\section{Learning Is Not Enough}
\label{sec:limits_of_learning}

Over the past decade rapid advances in model design and optimization have drastically improved our ability to train effective models in complex domains. Recent deep SL methods perform highly accurate modeling across modalities like natural language \citep{brown2020language, chowdhery2022palm,alayrac2022flamingo}, images \citep{karras2019style, ramesh2022hierarchical}, and speech \citep{oord2016wavenet, shen2018natural}. In parallel, progress in deep RL has produced agents matching or beating the top human players in strategic games like Go \citep{silver2016mastering, silver2017mastering} and Stratego \citep{perolat2022mastering}, while achieving state-of-the-art results in scientific pursuits such as chip design \citep{mirhoseini2021graph, roy2021prefixrl} and controlling nuclear fusion plasma \citep{degrave2022magnetic}. While impressive, these developments focus primarily on the learning problem---that is, the act of model design and optimization---and are a testament to the field's great progress in this endeavor. There remain many open questions on which aspects of intelligence cannot be adequately captured by existing DL approaches, e.g. robust symbolic reasoning \cite{marcus2020next}, causal understanding \citep{pearl2009causality, ortega2021shaking}, hierarchical planning in a world model \citep{lecun2022path}, and large-scale collective behaviors like cultural learning \citep{boyd2011cultural, henrich2015secret, ha2022collective}. We argue that, even if provided a joint model and optimization scheme that captures these missing aspects of intelligence, the training data itself remains a fundamental limitation and addressing it requires making progress on a host of separate problems. From this perspective, the historic focus on model design and optimization points to a systematic underinvestment in \emph{exploration}---how the data used to train these models is collected or created in the first place, whether in the form of a dataset or simulator. The generality of models trained in both supervised learning and reinforcement learning settings are ultimately bottlenecked on the diversity of this training data---a limitation demonstrated in several recent works \citep{cubuk2019autoaugment,cobbe2019quantifying,kaplan2020scaling,yarats2022don}. RL has historically focused on exploration as a principal objective for efficient learning, and indeed these ideas hint at a path toward addressing the data limitations of model learning in general. In this section, we consider these limitations in SL and RL and argue that despite its focus on exploration, RL ultimately suffers from the same data limitations as SL. We propose that we must rethink the existing exploration paradigm in RL to go beyond the data bottleneck on the path to general intelligence.

\subsection{The Limits of Supervised Learning}
Supervised learning aims to learn a function $f$ mapping points in domain $\mathcal{X}$ to domain $\mathcal{Y}$, given $N$ training examples $(x_i,y_i)_{i=0}^N$ reflecting this mapping, where $x \in \mathcal{X}$, $y \in \mathcal{Y}$. In modern ML, this function is typically a large artificial neural network with parameters denoted $\theta \in \Theta$. The defining feature of SL is that learning proceeds by optimizing a loss function $\mathcal{L}\colon\Theta\times\mathcal{X} \times \mathcal{Y} \to \mathbb{R}$ that provides the error between the model's prediction $f(x)$ and the true value $y$  paired with $x$.  In general, each training example $(x,y)$ can be seen as a sample from a ground-truth distribution $P(x,y)$, because the true data generating function $f^*\colon\mathcal{X}\to\mathcal{Y}$ can be stochastic. Therefore, the goal of SL can be seen as learning an approximator $f\colon\mathcal{X}\to\mathcal{Y}$ that produces samples $y$ consistent with $P(y|x)$, for example, by using a loss function that encourages $f$ to deterministically predict the mean or mode of $P(y|x)$, or that matches the distribution of outputs of $f(x)$ to $P(y|x)$.

Crucially, this definition of SL assumes the model $f$ is trained once on a static, finite training dataset, $\mathcal{D}_{\text{train}}$. We should thus not expect $f$ to accurately model data that differs significantly from its training data. Nevertheless, with massive amounts of training data, large models can exhibit impressive generality~\citep{brown2020language,chowdhery2022palm,ramesh2022hierarchical} and recent scaling laws suggest that test performance should improve further with even more data~\citep{kaplan2020scaling, hoffmann2022training}. Given the benefits of data scale, contemporary state-of-the-art SL models are trained on internet-scale, offline datasets, typically harvested via webcrawling. While such datasets may capture an impressive amount of information about the world, they inevitably fall short in containing all relevant information that a model may need when deployed in the wild. All finite, offline datasets share two key shortcomings: \textit{incompleteness}, as the set of all facts about the world is infinite~\citep{klein1999human}, and \textit{stationarity}, as such datasets are by definition fixed. For example, our virtual assistant, if trained on a static conversational corpus, would soon see its predictions grow irrelevant, as its model falls out of date with culture, world events, and even language usage itself. Indeed, all ML systems deployed in an open-world setting, with real users and peers, must continually explore and train on new data, or risk fading into irrelevance. What data should the system designer (or the system itself) collect next for further training? This is the complementary---and equally important---problem of exploration that sits below the surface of all ML systems in deployment, one that has been considered at length within the field of RL. 

\subsection{The Limits of Reinforcement Learning}
\label{subsec:limits_rl}
RL is a general formulation of the problem of online learning. In RL, the decision-making model, or \emph{agent}, interacts with its environment at each time $t$ by observing an observation $o_t$ emitted by the underlying Markov state $s_t$ of the environment and taking an action $a_t$ according to its \emph{policy} $\pi(a_t|o_t)$.  In response, the environment transitions to its next state $s_{t+1}$ according to the transition function $\mathcal{T}(s_{t+1}|s_t, a_t)$ and the agent receives a reward $r_{t+1}$ according to the reward function $\mathcal{R}(s_{t+1})$. Such an environment is called a Markov Decision Process (MDP). The policy $\pi$ is typically parameterized as a deep neural network with parameters $\theta$ and optimized to maximize the expected future discounted return---that is the discount sum of future rewards---assuming some prior $p$ over the initial state and a discount factor $\gamma \le 1$: 

\vspace{-5mm}
\begin{equation}
\label{equation:rl_objective}
J(\theta) = \mathbb{E}_{\pi} \left[\sum_{t} \gamma^t r_t \right].
\vspace{-1mm}
\end{equation}

The expected future discounted return from a specific state $s$  is also called the value of $s$. Here, the expectation over returns is over the specific distribution of states induced by following the policy $\pi$ in the environment. RL maximizes Equation~\ref{equation:rl_objective} by solving the \emph{credit assignment problem}, that is, estimating the future value of taking each action for each state of the world, thereby reducing the problem to a form of dynamic programming~\citep{bellman1966dynamic}: The optimal policy then consists of taking the highest value action in each state.

As RL agents train on their own experiences, locally optimal behaviors can easily self-reinforce, preventing the agent from reaching better optima. To avoid such outcomes and ensure sufficient coverage of possible MDP transitions during training, RL considers exploration a principal aim. Under exploration, the RL agent performs actions in order to maximize some measure of novelty of the resulting experiential data or uncertainty in outcome, rather than to maximize return. Simpler still, new and informative states can often be unlocked by injecting noise into the policy, e.g. by sporadically sampling actions uniformly at random. However, such random search strategies can run into the curse of dimensionality, making them often less sample efficient in practice.

A prominent limitation of state-of-the-art RL methods is their need for large amounts of data to learn optimal policies. This sample inefficiency is often attributable to the sparse-reward nature of many RL environments, where the agent only receives a reward signal upon performing some desired behavior. Even in a dense reward setting, the agent may likewise see sample-inefficient learning once trapped in a local optimum, as discovering pockets of higher reward can be akin to finding a similarly sparse signal. In complex real-world domains with large state spaces and highly branching trajectories, finding the optimal behavior may still require an astronomical number of environment interactions, even despite performing exploration. Thus for most tasks of interest, training an RL agent using real-world interactions is highly costly, if not completely infeasible. Moreover, a poorly-trained embodied agent acting in real-world environments can lead to potentially unsafe interactions. For these reasons RL is typically performed within a simulator, with which massive parallelization can reliably achieve billions of samples within a few hours of training~\citep{petrenko2020sample, freeman2021brax}. 

Simulation frees RL from the constraints of real-world training at the cost of the \emph{sim2real gap}, the difference between the experiences available in the simulator and those in reality. When the sim2real gap is high, RL agents perform poorly in the real world, despite succeeding in simulation. Importantly, a simulator that only implements a single task or small variations thereof will not produce agents that transfer to the countless tasks of interest for general intelligence. Thus, RL ultimately runs into a similar data limitation as in SL. Our virtual assistant, trained to navigate the web in a static simulation, cannot be expected to generalize to the many potential variants of webpages, both current and future, which may incorporate design, interaction, and structural paradigms completely missing in the original simulation. In fact, the situation may be orders of magnitude worse for RL, where unlike in SL \citep{kaplan2020scaling}, we have not witnessed results supporting a power-law scaling of test loss on new tasks, as a function of the amount of training data. Existing static RL simulators may thus impose a more severe data limitation on learning than large static, curated datasets, which has been shown capable of inducing strong generalization performance \citep{chan2022data}.

\subsection{Learning from Unlimited Data}
These fundamental limitations of both SL and RL result from their inherently offline formulation, whereby the training data is collected once from a blackbox process and provided to the learning process \emph{a priori}, in the form of a static dataset or predefined simulator. To keep its predictions relevant in an ever-changing world, a model must, instead, continually collect and train on new data---that is, it must perform continual exploration. We call such a data collection process \emph{generalized exploration} when it seeks to explore the full space of possible input data to the model. When this data space is unbounded, we may say the process performs \emph{open-ended exploration}, and a model trained on such a data stream then performs \emph{open-ended learning}. Though RL considers the problem of exploration at length, existing methods insufficiently address this form of data collection.

To see why this is the case, we first note that the MDP (i.e.~the simulator) plays an analogous role in RL to the training data in SL: Both define some distribution $\mathcal{D}$ over the space of available training data. In the case of standard SL, this distribution $\mathcal{D}$ consists of a uniform distribution over the individual data points $\{(x_i,y_i)\}_{i}$, where \mbox{$(x_i,y_i) \sim P(X,Y)$}. In RL, the simulator returns a training trajectory $\tau = (s_0, a_0, r_1, ..., s_T, a_T)$ for each policy $\pi$ and starting state, $s_0$. The distribution $\mathcal{D}$ is then the marginal distribution $P(\tau)$ under the joint distribution $P(\tau, \pi, s_0)$ induced by the MDP, some distribution over policies $\pi$, and the initial state distribution $P(s_0)$. In practice, the initialization of the policy network and RL algorithm determine the evolution of $\pi$ within the simulator, and thus both $P(\pi)$ and $P(\tau)$. We then see that exploration within the simulator is equivalent to sampling from $P(\tau | \pi, s_0)$ for a larger set of distinct $\pi$ and $s_0$ values, thereby increasing the support of $P(\tau)$, and consequently the number of distinct trajectories seen during training. Such a process can be viewed as a form of \emph{prioritized sampling}, in which data points favored by the exploration process---perhaps due to some form of novelty or uncertainty metric---are sampled with higher frequency. Applying standard RL exploration methods to SL then corresponds to actively sampling from $\mathcal{D}$. Importantly, such exploration methods do not expand the data available for training, but only the order and frequency in which the model experiences the existing data already within the support of a predefined, static distribution $\mathcal{D}$. For example, we can use such exploration algorithms to encourage our virtual assistant to uncover diverse and reward-rich trajectories during training, but the resulting data remain confined to the set of experiences represented by the static simulator, just as its conversational training data is limited to the exchanges captured in a static corpus.

Addressing the data limitation of SL and RL requires a different formulation of exploration, one in which the exploration process continually seeks new data to expand the support of the training distribution $\mathcal{D}$, so to provide the model with useful information---that is, where the model may experience the most learning. Achieving generality then requires the model to continually retrain on such newly acquired data---effectively learning on an unlimited stream of data at the frontier of its capabilities. Importantly, for RL, this entails discovering or inventing whole new MDPs, in which to collect transitions offering high information gain, e.g. those leading to the highest epistemic uncertainty as estimated by a Bayesian model \citep{williams2006gaussian, kendall2017uncertainties, clements2019estimating, hullermeier2021aleatoric}, regret relative to an oracle \citep{paired}, gradient magnitude, or gain in model complexity \citep{graves2017automated}. The question of what data should be collected for further training the model resembles that addressed by prioritized training approaches, including both active learning~\citep{settles2009active, cohn1994improving} and curriculum learning methods \cite{schaul2016prioritized,graves2017automated}, but differs in a subtle and important manner worth reiterating: prioritized training seeks to select the most informative data points for learning (and for which to request a ground-truth label in the case of active learning). Crucially, prioritized training assumes that the training data is otherwise provided \emph{a priori}. In contrast, we call our problem \emph{active collection}, whereby we must gather the most informative data for further training in the first place, and thus it is one that must precede any form of prioritized training. In defining active collection, we explicitly make clear the separation between \emph{optimization}, the process of fitting a model to data, and \emph{exploration}, the process of collecting this data. As we have argued, by training on static data and simulators, SL and RL implicitly presuppose that the problem of active collection has been solved, turning a blind eye to this critical enabler of learning. In accelerating the rate of information gain, active collection may be especially important for unlocking scaling laws for transfer performance to new tasks in RL, similar to those observed in SL.

In general, there are two ways to gather additional training data:

\textbf{Offline collection} occurs through a dedicated process separate from the model. For SL, this might entail using a web crawler to amass a comprehensive dataset of song lyrics, if it is found that the current assistant performs poorly in this domain. For RL, this might entail extending the simulator to include additional states, actions, transitions, and rewards that model important dynamics deemed missing. This process is equivalent to that used to collect the initial training data, with the exception that subsequent data collections can condition on the previous training data and current model performance in order to maximize some notion of information gain. Being largely driven by human expertise and ad hoc intervention, offline collection can be costly and susceptible to human biases. 

\textbf{Online collection} uses the model's own interactions with its deployment environment as additional training data. Any deployed model may be considered such an interactive ML system, which changes its environment with its own predictions~\citep{fails2003interactive}. In SL, this might entail simply retraining the model on a dataset of real-world interactions, e.g.~whether the user engaged with a recommendation or found a response to a query useful. In a purely RL setting, this might entail collecting trajectories in the deployment domain that are used to fine-tune or extend the simulator. In this case, the simulator may be a world model \citep{richalet1978model, hafner2020dreamer, hafner2021dreamerv2}---a DNN parameterization of the MDP that approximates the transition, reward, and observation functions---which can be retrained on this additional data. However, retraining on the model's own interactions can reinforce the model's existing biases.

Importantly, the potential biases of data collection can be mitigated by ensuring the data is \emph{actively} collected---that is, the collection process should seek datapoints maximizing some notion of potential information gain for the model. This collection can occur through a mix of manual guidance via a human expert and self-supervised approaches that directly generate or search for such highly informative datapoints. Active collection generalizes the form of exploration typically studied in RL, extending the domain of exploration beyond a single MDP and into that of an unbounded space of MDPs. By performing \emph{active offline collection}, we directly train the model on data that can effectively strengthen its known weaknesses and reduce the chances of collecting redundant data. Similarly, by performing \emph{active online collection}, we selectively sample only the most informative interactions for learning, which tend to be those that challenge the model's existing biases. Crucially, as we discuss in Section~\ref{sec:bootstrap_problem}, active collection avoids the highly problematic outcome of falling into local optima that can result from a ubiquitous bootstrap problem in online learning.

\section{The Bootstrap Problem in Online Learning}
\label{sec:bootstrap_problem}

\begin{figure}[t!]
    \centering
    \begin{subfigure}{0.33\textwidth}
    \includegraphics[width=0.8\linewidth]{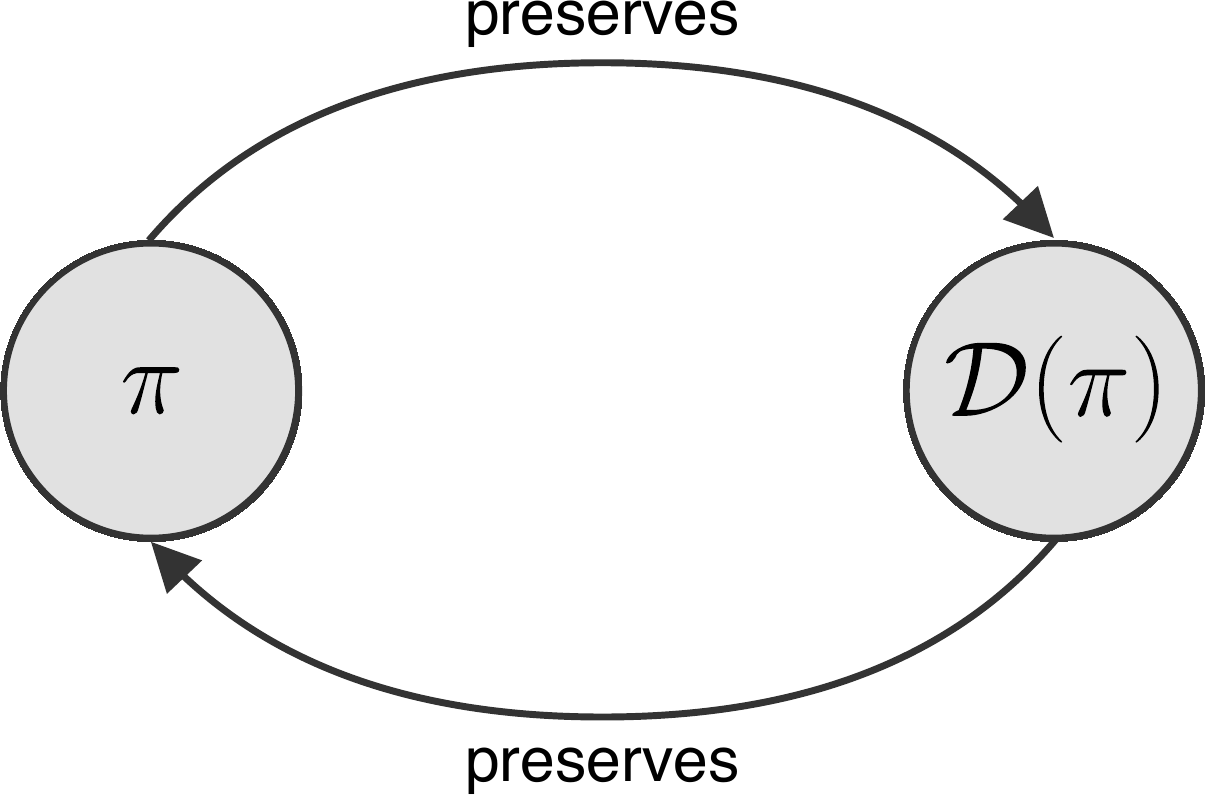}
    \end{subfigure}
    \vspace{3mm}
    \begin{subfigure}{0.33\textwidth}
    \centering\includegraphics[width=0.8\linewidth]{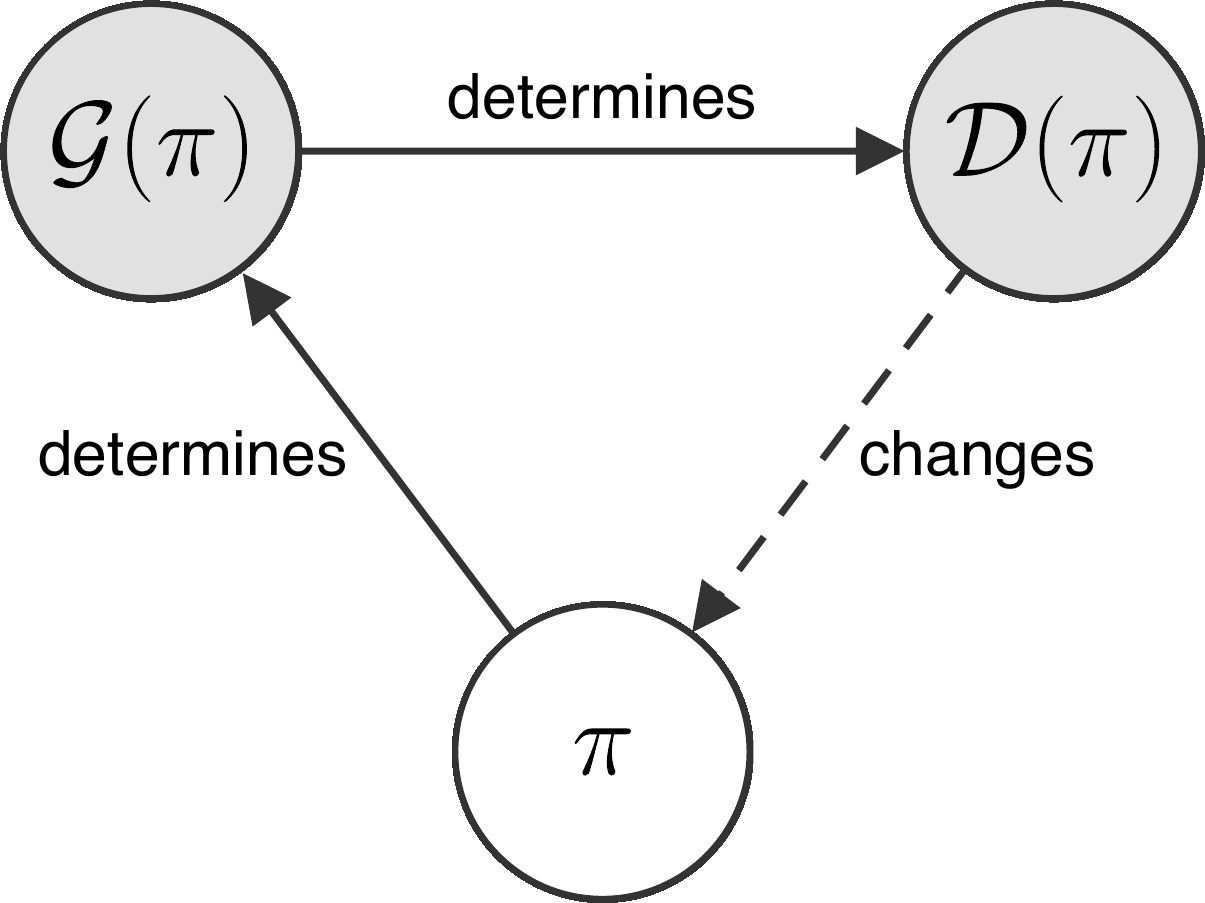}
    \end{subfigure}
    \caption{\small{Left: The bootstrap problem, due to performative prediction, leads to an equilibrium whereby the model fully determines a stationary data distribution $\mathcal{D}(\pi)$ that does not change upon retraining. As a result, both $\pi$ and $\mathcal{D}(\pi)$ become deadlocked in a fixed point. Right: Open-ended exploration delegates data collection to an active collection process $\mathcal{G}(\pi)$, which produces data $\mathcal{D}(\pi)$ designed to improve the current policy $\pi$ by presenting it with data on which its performance is weak. Under this system, $\pi$ continually improves, thereby avoiding fixed points.
    }}
    \label{fig:bootstrap_problem}
    \vspace{-4mm}
\end{figure}

Indeed, simply collecting data online in the deployment domain, without actively seeking it, is unlikely to produce an increasingly generally-intelligent model. More likely, the model will fall into a local optimum due to the particular dynamics of online learning: In online learning, the model retrains on newly acquired data, seeking to maximize an objective. Problematically, the data collected online is a function of the model's own predictions. For example, the state of the policy (e.g.~the model parameters, if it is directly parameterized) within an RL agent influences the experiences it will gain from interacting with the environment, thereby influencing the data from which the agent itself will be updated.
Left unaddressed, this causal loop between the data-generating process and the model's own predictions can result in a \emph{bootstrap problem}, where this feedback loop further amplifies any systematic errors or inherent biases in the model's predictions. 

Returning to our running example, we may decide to train our virtual assistant to shop for us on the web, where it consistently finds good deals on Amazon. As our assistant visits Amazon more often, it collects more online data of Amazon's webpages, which is used to refine the web simulation used for training. As the assistant becomes more adept at navigating Amazon, it consistently attains high rewards when fulfilling our requests there, further reinforcing this preference, which then grows to the extent that it never visits any other websites, where there may be much better deals. Iterated retraining on data collected online thus causes our agent to become stuck in a local optimum. This bootstrap problem, depicted in Figure~\ref{fig:bootstrap_problem}, impacts all online learning systems. It has been studied across various problem settings, including in contextual bandits, where it goes by the name of \emph{one-sided learning}~\citep{helmbold1992apple, jiang2021learning}, and classification, under the game-theoretic lens of \emph{strategic classification}~\citep{hardt2016strategic}. 

This bootstrap problem has been formalized in the problem setting of \emph{performative prediction}~\citep{perdomo2020performative}. Here, the model $f$ induces a distribution $\mathcal{D(\theta)}$ over the training data $Z$. The problem arises from the equilibria of this dynamic system, between model and data distribution. At such fixed points, the data distribution induced by the model, with parameters $\theta^*$, is identical to that on which it was last retrained and which induced the solution $\theta^*$. With these definitions, we can define the problematic points $\theta^*$ in the model's parameter space as those satisfying

\begin{equation}
\theta^* = \underset{\theta}{\arg\min}\;\mathbb{E}_{Z \sim \mathcal{D}(\theta^*)}[\mathcal{L}(Z,\theta)].
\end{equation}

While prior analysis focuses on the convergence toward such fixed points and on their optimality, these are exactly the points that agents seeking to learn general capabilities must avoid: At such fixed points, the online learning system stagnates into a subset of training data, preventing the possibility of learning anything beyond the current data distribution. Despite the bootstrap problem, online retraining is commonly implemented in production systems \citep{bellogin2017statistical, ge2020understanding, neophytou2022revisiting, karpathy2021cvpr, karpathy2021tesla}, while this crucial issue remains largely unaddressed, resulting in unfavorable emergent phenomena, such as extremization through personalized content recommendation systems and the amplification of socioeconomic disparities through AI-powered lending systems \citep{bartlett2022consumer, hong2016political, tucker2018social, bak2021stewardship}.

Standard exploration techniques in RL and prioritized training techniques in SL have been proposed to address this issue. Exploration and prioritized training both select the next training data points that are, by some criterion, those on which the model is most likely to be wrong, thereby forcing the model to reconsider its decisions and thus deviate from any premature equilibrium. In this sense, prioritized sampling could be described as a form of exploration, and conversely, exploration in RL, as a form of prioritized sampling. Such methods may suffice to avoid local optimum due to the bootstrap problem in a limited-data setting, though of course, reaching the globally optimal solution still ultimately corresponds to falling into a fixed point, beyond which no further learning occurs. The central, limiting factor is that existing exploration methods target a limited data setting, operating on either a fixed dataset or a static simulator. In order to go fully beyond exploratory equilibria, we must rethink exploration to encompass exploring beyond any one dataset or simulator.

\section{Open-Ended Exploration}
\label{sec:exploring_beyond}

How can we perform exploration beyond a single static dataset or simulator? Before tackling this challenge, it is valuable to consider how exploration is currently performed in static, limited-data settings. Exploration in RL typically aims to maximize some criterion of novelty or uncertainty~\citep{amin2021survey}. When the same evaluation criterion is used to select the next datapoints for training in SL, the process reduces to prioritized training. In this section, we center our discussion on exploration in RL for a static simulator, then consider how such strategies can translate to supervised learning.

\subsection{Exploration in a Static Environment}
Let us call the RL policy that seeks to maximize the discounted return in a given MDP the \emph{exploitation policy}. As discussed, the exploitation policy can become trapped in local optima, due to the bootstrap problem. Exploration seeks to address this issue by producing a second policy, which we call the \emph{exploration policy}, to collect training data for the exploitation policy that is unlikely to be generated by simply following the exploitation policy. 
In order to find such data, the exploration policy typically either performs random actions in each state, or seeks to maximize a measure of novelty---that is, transition data that looks different from what it has previously experienced, which may be computed on an episodic or lifelong basis. A common measure of novelty for state $s$ is based methods for counting state visitations, whereby less visited states are consider more novel~\citep{ostrovski2017count}. In practice, there are many proxy measures of novelty, like epistemic uncertainty, which may be estimated by the variance of a prediction network~\citep{pathak2019self, sekar2020planning}. Similarly, prediction error \citep{oudeyer2007intrinsic, kompella2017continual} or regret---the difference between optimal return and that achieved by the agent---can also be used as measures of novelty, assuming the agent fares worse on environments that present novel challenges. Importantly, when applied to the mean performance over a batch of trajectories, these latter approaches implicitly subtract away sources of inherent, irreducible uncertainty \citep{kendall2017uncertainties,der2009aleatory} that can act as \emph{stochastic traps} for a novelty-seeking agent. Alternatively, such irreducible uncertainty can be removed from the calculation by explicitly modeling it \citep{mavor2022stay}. In general, the exploration policy aims to maximize an \emph{intrinsic reward function}, $I:S\times A \rightarrow \mathbb{R}$ \citep{singh2010intrinsically}. 

Beyond the shared motivation for seeking informative transitions, methods differ in how exploration is folded into training. One common tactic is to maintain a separate exploration policy, e.g.~one that learns to maximize the future novelty of the exploitation policy. In this setting, the exploration policy is typically called the \emph{behavior policy}, as it is solely responsible for data collection, while the exploitation policy is called the \emph{target policy}. The target policy then trains on the transitions collected under the behavior policy using importance sampling \citep{sutton2018reinforcement, munos2016safe}. Another increasingly popular approach is to use a single policy to serve as both exploration and exploitation policies \citep{pathak2017curiosity,burda2019exploration, raileanu2020ride, zhang2021noveld, henaff2022exploration}. This single policy takes actions to maximize a weighted sum of extrinsic and intrinsic returns. As exploration continues to reduce the uncertainty in most states of the MDP, the intrinsic reward tends to zero, resulting in an approximately purely exploiting policy at convergence, though in general, annealing the intrinsic term may be required \citep{singh2000convergence}. A related set of approaches based on \emph{probability matching} samples actions according to the probability that each action is optimal, where actions may receive some minimum amount of support to encourage exploration, or support that increases with the estimated uncertainty of the transition resulting from taking the action~\citep{thompson1933likelihood, osband2016deep}.

\subsection{Exploration in Environment Space}

Our discussion of exploration in a single, static MDP suggests close parallels to the more general problem of active collection, with the major difference being that exploration methods in RL focus on a single, predefined MDP---as defined by the static simulator---rather than over the infinite set of all potential MDPs, as sought by active collection. Exploration in the space of parameterized environments offers a path towards this more general form of exploration.\footnote{The set of all MDPs can in general be indexed into a single \emph{universal MDP} by assigning each a unique real value index and including this index in the state. Exploration over the space of MDPs can then equivalently be viewed as exploring this universal MDP. We instead frame the discussion in terms of the space of MDPs to emphasize and exploit the inherent modularity of the problem.}

\subsubsection{Exploring a Subspace of Environments}
\label{subsubsec:explore_env_subspace}

In recent years, RL research has reoriented around learning optimal policies for distributions of environments \citep{risi2020increasing,cobbe2019quantifying,kuttler2020nethack}. In this expanded setting, the environment is parameterized by a set of \emph{free parameters}, whose values alter properties such as the environment topology, entities present, transition dynamics, and the underlying reward function. The choice of free parameters specifies a particular subspace of environments, within the space of all possible environments. Simply randomizing over these parameters during training, in what is called \emph{domain randomization}, can lead to robust policies. Automatic curriculum learning methods~\citep[ACL;][]{portelas2020automatic}, which produce adaptive curricula over these parameters have been shown to further improve robustness. Such adaptive curricula adjust the free parameters of the environment throughout training to maximize a measure of the agent's learning potential, resulting in an adaptive curriculum over specific environment configurations. Recently, unsupervised environment design~\citep[UED;][]{paired} extends the problem setting of \emph{decision-making under ignorance} \citep{peterson2017introduction} to capture the notion of robustness and frames the problem such that solutions naturally take the form of adaptive curricula. In UED, the MDP tuple is augmented with an additional set of free parameters, denoted $\Theta$, whereby possible environment configurations correspond to specific settings $\theta$ of these parameters. An adaptive curriculum then arises from a game between a teacher, which proposes configurations $\theta$, and a student that learns to solve them. The choices of payoffs for the student and teacher then correspond to an infinite spectrum of possible adaptive curricula. For example, DR corresponds to a constant payoff for the teacher---so that it is indifferent to the value of $\theta$, and thus randomizes over it. When the game is zero-sum, with the teacher receiving the regret of the agent as its payoff, any Nash equilibrium \citep{nash1950equilibrium} corresponds to the student playing a policy that minimizes the worst-case regret---that is, the minimax regret policy~\citep{savage1951theory}.

Curriculum games, in which the teacher seeks to generate environment configurations where it is possible for the agent to rapidly improve, lead to co-evolutionary dynamics that gradually increase the difficulty of these configurations, as the student grows more adept. Importantly, the teacher's payoff $C:\Theta \rightarrow \mathbb{R}$, should adapt to the agent's capabilities by satisfying these criteria for high learning potential:

\begin{enumerate}
    \item \textbf{Improvability:} The agent does not fully succeed. There is room for improvement.
    \item \textbf{Learnability:} The agent can efficiently learn to improve.
    \item \textbf{Consistency:} The solution to each environment configuration is consistent with those of other configurations.
\end{enumerate}

If the proposed environment configurations do not satisfy (1), then the configuration provides no additional benefit for training, as the current model already performs optimally. If the configuration does not satisfy (2), then despite there being room for improvement, the current model is unlikely to improve when training on that configuration. Presenting many configurations that fail to meet the learnability criterion can stall learning progress, as the agent struggles to learn the solution. Finally, (3) ensures that optimal behavior in a proposed configuration does not conflict with that in another, e.g.~entities are semantically consistent. As in exploration in a singleton MDP, learning potential estimates must remove sources of aleatoric uncertainty to avoid stochastic traps. Such corrections can be done by directly modeling and subtracting sources of aleatoric uncertainty or estimating mean learning potential over a batch of trajectories for each configuration. 

Just as exploration methods in a singleton MDP encourage the agent to visit novel parts of the state space to maximize opportunities for learning, adaptive curricula bring the agent to the environment configurations that offer the most learning potential, itself a form of novelty. In this way, adaptive curricula perform exploration in a parameterized space of environments. 

\subsubsection{Exploring the Full Space of Environments}
\label{subsubsec:explore_env_full_space}
Still, a parameterized environment only allows agents to learn to succeed in the specific configurations expressible within its parameterization. Since the parameters only modify the instantiation of a static simulator designed to model a limited task domain, the agent's learning will eventually plateau. Open-ended exploration requires an open-ended simulation, one that continually expands to encompass new domains---in essence, exploring the space of possible MDPs. To this end, we can define a search process $\mathcal{G}(\pi)$ that takes the current policy $\pi$ and returns a distribution $\mathcal{D}_{\mathcal{M}}$ over the set of MDPs, i.e.~a countably infinite set of programs under some Turing-complete language. This distribution aims to focus its support over MDPs that maximize the learning potential criterion, $C:\mathcal{M}\times \Pi\rightarrow\mathbb{R}$. By evaluating $\pi$ on $m \sim \mathcal{G}(\pi)$, we can iteratively update $\mathcal{G}(\pi)$ toward  distributions over $\mathcal{M}$ that place greater weight on MDPs with higher learning potential. In practice, $\mathcal{G}$ can consist of large parametric models alongside non-parametric models and human-in-the-loop components to assist with the search process. Moreover, the exact components implementing $\mathcal{G}$ may shift over time. The search process $\mathcal{G}$ then seeks MDPs \mbox{$m^* \in \mathcal{M}$} maximizing the exploration criterion:

\vspace{-2mm}
\begin{equation}
    \label{eq:oel_exploration_criterion}
    m^* = \underset{m \in \mathcal{M}}{\arg\max}\;C(m, \pi).
\end{equation}

In practice, we might model $\mathcal{G}$ as a large generative model, whose outputs are MDP programs, and its weights, fine-tuned to maximize the exploration criterion, $C$. Left alone, this approach runs into three interrelated issues: First, the programs generated will be largely malformed, and an astronomical number of iterations may be necessary to discover the first valid MDP program. Second, the programs discovered will likely be nonsensical, failing to reflect the target problems of interest. Third, upon discovering valid programs, exploration may overly focus on a limited subset of the full space of programs. In order to generate both well-formed and relevant MDPs, we must constrain the search to begin near well-formed programs reflecting the target tasks of interest. We use the phrase \emph{grounding with respect to} $\overline{\mathcal{M}}$ to refer to the process of enforcing this constraint with respect to a target set of seed MDPs, $\overline{\mathcal{M}} = \{m_k\}_{k=1}^K$. In order to maintain diversity in the programs explored, we can directly encourage diversity in the space of MDPs. Given distance functions $\Delta_D$ and $\Delta_G$ between two programs, we can ground open-ended exploration with respect to $\overline{\mathcal{M}}$ and encourage diversity by augmenting the exploration criterion in Equation~\ref{eq:oel_exploration_criterion}:

\begin{equation}
    \label{eq:oel_exploration_criterion_grounded}
    m^* = \argmaxbigg{m \in \mathcal{M}}{\!\!\!\!\!\!\!\!\!\underbrace{\vphantom{\sum_{m_i \in \hat{\mathcal{M}}}}{C(m, \pi)}}_{learning\;potential} \!\!\!\!\!\!+\;\;\underbrace{\alpha_D\!\!\!\sum_{m_i \in \hat{\mathcal{M}}} \Delta_{D}(m, m_i)}_{diversity}\;\; - \underbrace{\sum_{m_k \in \overline{\mathcal{M}}}\alpha_k\cdot\Delta_{G}(m,m_k)}_{grounding}},
\end{equation}

\begin{figure}[t!]
    \centering
    \begin{subfigure}[b]{0.25\textwidth}
        \includegraphics[width=1\textwidth]{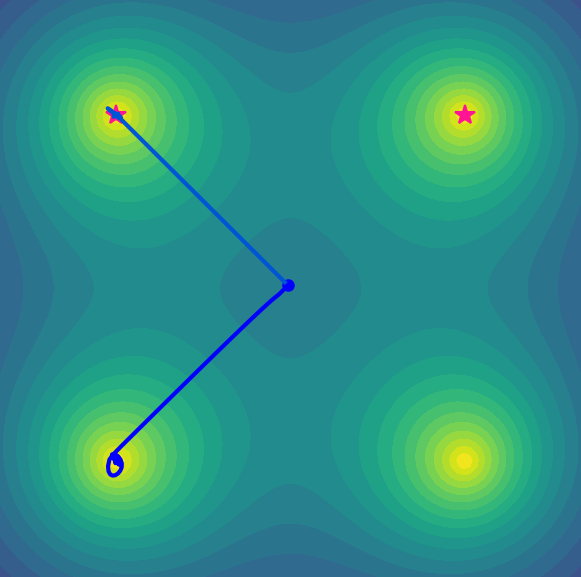}
    \end{subfigure}
    \hspace{0.025\linewidth}
    \begin{subfigure}[b]{0.25\textwidth}
        \includegraphics[width=1\textwidth]{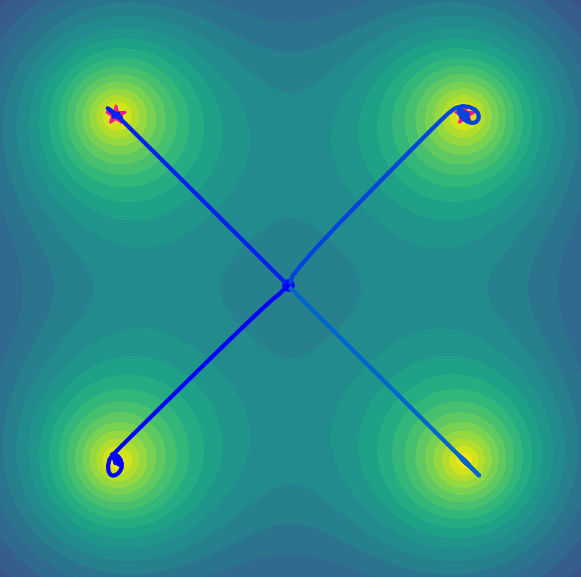}
    \end{subfigure}
    \hspace{0.025\linewidth}
    \begin{subfigure}[b]{0.25\textwidth}
        \includegraphics[width=1\textwidth]{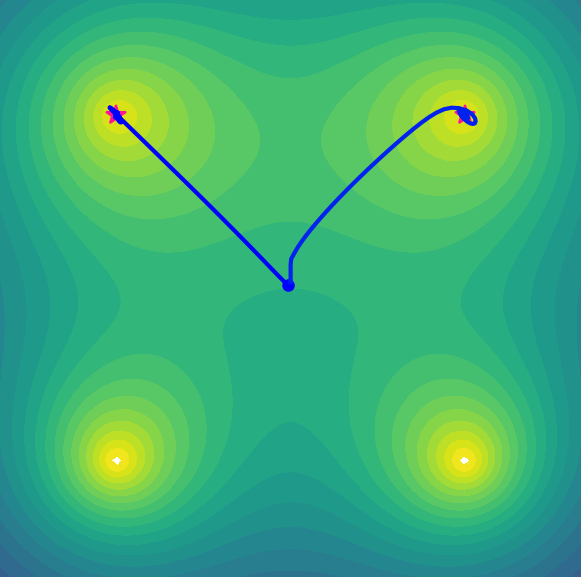}
    \end{subfigure}
    \caption{\small{A simple visualization of how an open-ended exploration process searches for points in task space according to Equation~\ref{eq:oel_exploration_criterion}. Brighter regions have greater learning potential. The pink stars are tasks of interest, and the blue lines are search paths. Left: Searching only based on learning potential can lead to both useful tasks and irrelevant tasks. Middle: Searching with an additional diversity term can lead to finding both more relevant and irrelevant tasks. Right: Grounding the search to useful tasks leads to finding more relevant tasks.
    }}
    \label{fig:oe_exploration_criterion_visual}
    \vspace{-6mm}
\end{figure}

where $\alpha_D$ and $\alpha_k$ are positive weights modulating the relative importance of the diversity bonus and being close to $m_k \in \overline{\mathcal{M}}$ respectively, and $\hat{\mathcal{M}}$ is finite queue of the top solutions found so far. We explicitly define two distinct distance functions as, in general, the diversity and grounding terms might consider different notions of distance, e.g. the string edit distance or a distance in the latent space representation of a pretrained generative model of programs. When computed with respect to specific attributes of generated programs, $\Delta_G$ may encode inductive biases of what useful MDPs look like, e.g.~whether a given MDP simulating a physical system produces trajectories consistent with the laws of physics. When humans are part of the open-ended exploration process, they can regularly modify the set of seed MDPs, $\overline{\mathcal{M}}$, used for grounding, and steer exploration toward specific regions in the space of MDPs deemed important by adjusting the corresponding $\alpha_k$ weights. Figure~\ref{fig:oe_exploration_criterion_visual} provides a toy illustration of how these criteria impact the search process.

It is important to draw a similarity between this exploration criterion and the objective of algorithms for quality-diversity~\cite[QD;][]{pugh2016quality}, commonly explored in the evolutionary computing community. Such algorithms seek to find a diverse set of solutions, where the best candidates within distinct subspaces of the solution space are gradually discovered. Viewing the learning potential term as a measure of quality, we see that Equation~\ref{eq:oel_exploration_criterion_grounded} implements a form of QD. However, this process remains significantly distinct from QD, as the measure of learning potential is a function of the current model, and therefore non-stationary throughout training. Moreover, learning potential is not transferable as a quality measure across agents in different training runs. To acknowledge the similarities to QD while highlighting its distinctness from QD, we call the problem of finding diverse data with high-learning potential that of \emph{learning-diversity}.

Similar to the outlined open-ended exploration process, prior work on the POWERPLAY algorithm has considered finding increasingly general agents by jointly searching through the space of MDP programs and that of solution programs \citep{schmidhuber2013powerplay}. We detail the relationship between our proposed approach to open-ended learning and POWERPLAY in Section~\ref{sec:bottom_up_agi}. Importantly, POWERPLAY frames the problem purely as a form of iterated optimization via random search, ultimately a computationally infeasible approach. In contrast, Equation~\ref{eq:oel_exploration_criterion_grounded} defines an exploration criterion allowing for agents to gradually increase their generality by actively collecting and training on data from diverse regions of the MDP space with high learning potential, while remaining grounded and steerable by expert guidance, e.g.~by a human. Importantly, this generalized form of exploration not only applies to RL in MDPs, but can also be straightforwardly translated to SL. 

\subsection{Exploration in Data Space}

In viewing a static SL dataset as a parameterized, single-step MDP, we can similarly implement open-ended exploration by pursuing the criterion in Equation~\ref{eq:oel_exploration_criterion_grounded}. For a static dataset $\mathcal{D} = \{x_i,y_i\}_{i=1}^{N}$, where $x_i \in \mathcal{X}$ and $y_i \in \mathcal{Y}$, we can  reframe the problem as finding the optimal policy for an MDP $m$, whose optimal policy $f$ is the model that minimizes the empirical risk on $\mathcal{D}$. In this case, $m = (S,A,T,R,\gamma,p)$, where the set of states is defined as $S = \{x_i\}_{i=1}^N$, the set of actions $A = \mathcal{Y}$, the transition function $T$ simply terminates the episode, and the reward function $R$ returns the negated supervised loss between the predicted value $f(x)$ and target value $y$. Since the MDP is single-step, $\gamma$ can be any value in $[0,1]$. The initial state distribution $p$ is a uniform distribution over $S$. As defined, the optimal policy $f$ for $m$ must minimize $\mathbb{E}_{x_i\sim p}\left[\mathcal{L}(f(x_i), y_i)\right] = \frac{1}{N}\sum_{i=1}^N \mathcal{L}(f(x_i), y_i)$, exactly the empirical risk over $\mathcal{D}$, as typically minimized under SL. 

Having established the correspondence between SL datasets and single-step MDPs, we see that open-ended exploration in SL entails active collection of datasets that maximize the exploration criterion outlined in Equation~\ref{eq:oel_exploration_criterion_grounded}. This view makes clear that sampling strategies used in prioritized training implement specific instantiations of this criterion, limited to a fixed training dataset, $\mathcal{D_{\text{train}}}$. 
For instance, active learning selection functions, as well as those in curriculum learning, are largely based on choosing the next training sample to maximize the information-gain, or relatedly to reduce some notion of uncertainty over the model's own predictions or variance in performance error \citep{settles2009active,houlsby2011bayesian}. Any such selection function corresponds to a specific choice of the learning potential subcriterion, $C(m, \pi)$, where $m$ now corresponds to a datapoint in $\mathcal{D_{\text{train}}}$, and $\pi$, the supervised learning model. Curriculum-based prioritized training largely mimics the same selection strategies \citep{mindermann2022prioritized, schaul2016prioritized, graves2017automated}, while removing the interactive label queries of active learning. 

This framing also points to the rich possibilities that lie in extending such prioritized training methods toward the full scope of open-ended exploration: New variants of prioritized training may benefit from additionally incorporating approximations of the diversity and grounding subcriteria. The diversity term helps prevent oversampling of particularly challenging points at the expense of other informative points. The grounding term comes into play when incorporating various kinds of generative procedures based on the training data to perform data augmentation, ensuring generated samples do not go too far afield from the real data. Early results suggest such generative models hold great promise for improving the robustness of SL models when used to generate \emph{synthetic training data} \citep{zhang2017mixup, jahanian2021generative}. Such data generation may be particularly effective for amplifying the amount of available data for rare samples, e.g. extreme weather conditions or unique vehicular appearances and interactions captured by camera feeds for training the models underlying a self-driving car.

Importantly, by generating \emph{synthetic data} grounded to real task domains, the exploration process can produce alternate views of existing data that serve as interventions of specific causal features within the data space \citep{von2021self, ilse2021selecting, mao2021generative}, which can lead to models whose predictions are more consistent with the underlying causal structure of the task domain \citep{pearl2009causality}. As data generators, such generative models play an analogous role to the simulator or world model in RL, providing alternative views, and allowing the model---guided by an appropriate exploration critrion---to probe the data space through trial and error or more sophisticated experimentative behaviors. As in the case of RL, online data collected in deployment can be used to further fine-tune the generative models to continually expand the space of represented data. More generally, as in RL, we can pursue the open-ended exploration criterion in SL through a combination of curating useful interaction data (i.e.~online collection), manually collected seed datasets that reflect our domains of interest (i.e.~offline collection), and actively generating data with high learning potential for our model using powerful generative models such as state-of-the-art diffusion models~\citep{sohl2015deep, song2021maximum, kingma2021variational, nichol2022glide}, grounded to these seed datasets. 

\section{A Unified View of Exploration}
\label{sec:unified_view}

Our discussion reveals that exploration can be reconceptualized as a search process over the space of MDP programs. Such a view of exploration highlights direct correspondences between prioritized training methods in SL and exploration in RL, while radically expanding the domain of exploration to include the full space of possible training data. Crucially, exploration remains the same process of finding the most informative datapoints for learning, regardless of the the optimization procedure used---whether based on an SL or RL objective. Open-ended exploration performs this search over the full data space and  serves as the core driver of information gain in open-ended learning systems seeking general capabilities. We now ground these ideas to a concrete example, illustrating how such open-ended exploration might be implemented in practice to give rise to a learning system of increasingly general intelligence. The differences between our hypothetical open-ended learner and ad hoc forms of exploration used in practice today then reveal a new data-centric research frontier.

\subsection{A Unified Example}
\begin{figure}[t!]
    \centering
    \includegraphics[width=0.9\textwidth]{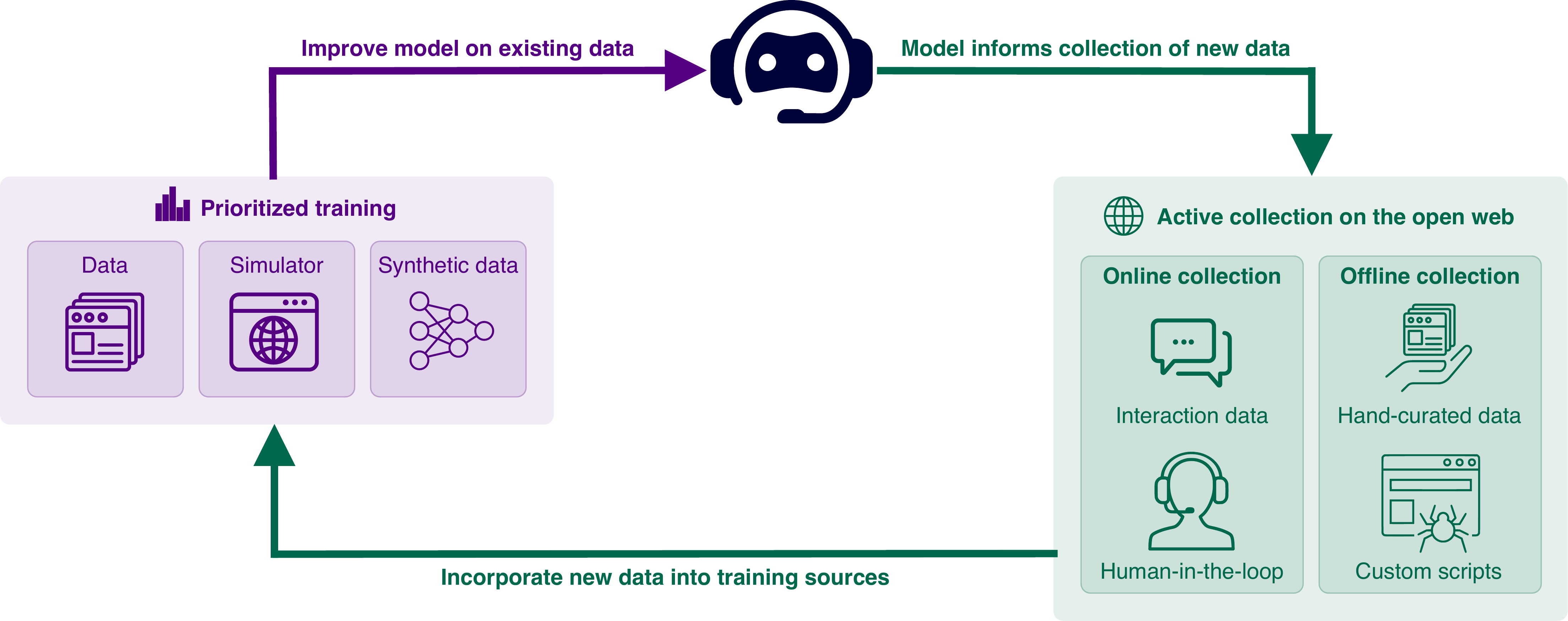}
    \caption{\small{An overview of how active collection and prioritized training combine to enable open-ended exploration and learning to produce a virtual assistant with increasingly general capabilities. An initially trained assistant model is used to inform the active collection of new data from both offline and offline sources, possibly with assistance from humans and other programs in the loop. The newly collected data is then used to update the previous training sources, which might include labeled datasets, simulators, and synthetic data, e.g. sampled from generative models or a learned world model.}}
    \label{fig:oe_virtual_assistant}
    \vspace{-4mm}
\end{figure}

Let us think through how a concrete system might perform such open-ended learning by returning to our example of a virtual assistant. Recall that our virtual assistant must perform two key functions: (1) converse with a user via text, and (2) navigate any website to accomplish tasks on behalf of the user. Either (1) or (2) in itself already approaches the difficulties of training a generally intelligent agent; both language and software offer universal representations of any computable task, so becoming generally competent in either domain entails achieving a form of general intelligence---the ability to successfully handle any computable task. Of course, the space of all computable tasks (i.e.~$\mathcal{M}$) is vast and even humans are not generally intelligent in this regard. Nevertheless we can pursue continual incremental progress toward this goal by training an IGI through open-ended learning. Such a system, summarized in Figure~\ref{fig:oe_virtual_assistant}, performs three distinct, interlocked modes of learning.

\textbf{Initial training} As usual, we design an initial dataset to seed the training of our virtual assistant. As done in the training regimes of current state-of-the-art conversational models, this dataset likely includes a large corpus of dialogue mined from conversations between real people. As performed in recent works, RL can be used in combination with a human-designed reward function, i.e.~reinforcement learning with human feedback~\citep{christiano2017deep,ouyang2022training}, to encourage responses that are both consistently friendly and helpful. Following the example of other grounded interaction models like LaMDA~\cite{thoppilan2022lamda}, web navigation agents can similarly be bootstrapped from a combination of SL over a dataset, perhaps collected from trajectories generated by hand-coded controllers for performing common tasks on a diverse set of webpages, and RL inside of a web navigation simulator~\citep{shi2017world, gur2021environment,nakano2021webgpt}. To connect its conversational and web navigation functions, synthetic and augmented conversation data, both generated offline and within a simulator, can serve as templates for successful dialogues whose responses require web navigation.

\textbf{Active collection} After initial training, we can deploy our assistant to users. At deployment, we perform active online collection, curating a dataset of interactions in which the assistant performed poorly, according to some proxy metric such as whether users engaged with the content shared by the assistant. In parallel, active offline collection can be performed, whereby specific weaknesses in the model are identified, e.g.~websites where navigation commonly fails or semantic clusters of request types that result in low engagement. These areas are then targeted for the collection of real dialogue data and expert web-navigation trajectories, resulting in an \emph{adversarial} dataset. This collection can occur through many channels, e.g.~manually through crowdsourced efforts; storing the interactions resulting from a human-in-the-loop taking over for the assistant when the assistant's predictions become highly uncertain; and custom webcrawlers that harvest relevant conversational data or videos from YouTube that serve as expert demonstrations \citep{baker2022video}. The new states, actions, and dynamics present in this data can further be integrated into the existing simulator, either through fine-tuning if the simulator is a learnable model, e.g.~a world model, or through manual coding. Further, this data can be used to train a generative model, grounded to the real data, thereby greatly amplifying the amount of data and allowing for fine-grained exploration of the relevant regions of the data space. The assistant can then be retrained on this adversarial data to become more robust. This active collection and retraining cycle then repeats, resulting in an agent that becomes robustly capable across an increasing number of domains. Importantly, we see that exploration naturally occurs over the data space itself---not within some notion of a static simulator or dataset. This example also highlights the blurring of the lines between simulated and real data. Real data is used to generate further simulated data, and performance on simulated data informs the active collection of real data.

\textbf{Prioritized training} The assistant performs prioritized training to efficiently learn from the collected data. In SL, there exist many prioritized training methods~\citep{kirsch2019batchbald, graves2017automated} that can be used to selectively sample datapoints that offer the most learning potential. In RL, methods like Prioritized Experience Replay~\citep{schaul2016prioritized} serve the same purpose. Other, more recent \emph{unsupervised environment design} \citep{paired, gur2021environment, robustplr, accel} approaches mix active collection and prioritized training, whereby the algorithm actively searches for new MDPs, e.g.~new website configurations, that offer the most promising environment configurations for learning, and then actively samples these configurations during training. Viewing a generative model as a parameterized dataset, we can apply the same approaches to both explore the data space of the generative model and actively sample the discovered points for training.

\vspace{-2mm}
\subsection{A View Beyond the Frontier}
Many real-world systems, including those controlling large-scale applications such as video recommendation \citep{baluja2008video} and traffic prediction \citep{hu2022deepreta}, use ad hoc versions of iterated retraining, relying on online and offline collection processes that are largely agnostic to learning potential. As such, their online collection can suffer from the bootstrap problem, falling into premature equilibria. Similarly, their offline collection may lack sufficient exploration of the data space. While we do not offer a specific solution for effectively pursuing the open-ended exploration criterion in Equation~\ref{eq:oel_exploration_criterion_grounded}, we believe articulating these criteria can be useful in motivating future work in this area.

As our state-of-the-art learning methods approach asymptotically-optimal performance for static training sets and benchmarks, the primary lever for improving our models lies in improving the training data itself. 
Several results support this data-centric view of ML: For large neural models, continually scaling up the quantity of training data empirically improves generalization performance~\citep{nakkiran2021deep,kaplan2020scaling,hoffmann2022training}. Generating additional supervised samples through methods like \emph{data programming}---which generates new labels based on human-provided, approximate labeling functions, results in significant performance gains in several domains~\citep{ratner2016data,ratner2017snorkel}.

We foresee the pendulum of research will swing from \emph{learning innovation} in the form of better model architectures and optimizers for training on specific data to \emph{data innovation} in the form of more intelligent algorithms for collecting additional data for improving a specific model. This turn toward data innovation is the natural conclusion of the paradigm shift in software introduced by the advent of deep learning. This shift transitioned software development from the largely imperative approach of directly designing solutions in code to the declarative approach of simply designing an objective function over training data that is optimized by a DNN---an approach sometimes referred to as \emph{Software 2.0}~\citep{karpathy2017software2,carbin2019overparameterization}. The solution then emerges as a result of this optimization, but the choice of training data remains a highly manual design problem. The data innovation paradigm then goes one step further, by also replacing the direct design of training data with the design of the exploration criterion, that is, a self-adapting data-generation process. We may thus refer to this data-centric paradigm as a kind of \emph{Software Squared}. Like the shift from designing solutions to designing objectives, we believe the shift to designing the data-generation process itself will change the face of computing and enable a new generation of more generally intelligent software.

\section{Intelligence as Capacity for Exploration}
\label{sec:bottom_up_agi}

Open-ended exploration presents a promising path to the longstanding goal in AI research of developing increasingly more powerful Artificial General Intelligence (AGI)---that is, systems capable of matching or exceeding humans in an ever wider range of tasks. The very act of defining a general notion of intelligence in humans and animals has been historically fraught with debate. Unsurprisingly, the quest to produce an agent implementing general intelligence is no less contentious. Many alternatives toward this grand goal have been proposed. Here, we relate the open-ended path to general intelligence to the most relevant alternative paths. 

The precise definition of AGI is ultimately a subjective one, based on the specific criteria and assumptions held. As previously mentioned, we take the view that general intelligence cannot be quantified in degree, as such a computation is necessarily intractable given the countably infinite size of the space of possible, computable tasks. Rather, we can only compare the relative degree of generality held by two models $A$ and $B$, with respect to a specific set of tasks, $\mathcal{T}$. We can then say $A$ is more general than $B$ in $\mathcal{T}$, if $A$ performs above a performance threshold on more tasks in $\mathcal{T}$ than $B$, such that $A$ also at least matches the performance of $B$ on all tasks that $B$ performs above the threshold.\footnote{Naturally, this definition implies that there may be no single most general agent in some domains, such as zero-sum games, for which the ordering of strategies is not guaranteed to be transitive \citep{czarnecki2020real}.} Under this definition, a learning process may hold the capacity to become increasingly general over time, by incrementally mastering a wider range of challenges. Exploration, as we have argued, serves as the central mechanism enabling this gradual expansion of capabilities. 

AIXI \citep{hutter2007universal}, perhaps the most prominent model of AGI, runs counter to our viewpoint. AIXI is the globally optimal solution to the Universal Measure of Intelligence, $\Upsilon$ \citep{legg2007universal}, defined in~\ref{eq:universal_intelligence_measure}.

\begin{equation}
    \label{eq:universal_intelligence_measure}
    \Upsilon :=  \sum_{\mu \in \mathcal{M}} 2^{-K(u)}V^{\pi}_{\mu}
\end{equation}

This measure, $\Upsilon$, defines the degree of general intelligence for a decision-making policy $\pi$ as the sum of the expected performance of $\pi$ in every computable MDP $\mu \in \mathcal{M}$, as captured by its value function, $V_{\mu}^{\pi}$. Crucially, the contribution of each MDP $\mu$ is inversely weighted according to its Kolmogorov complexity, $K(\mu)$, the length of the shortest program, represented as a bitstring, that implements $\mu$. A shortcoming of this definition is that Kolmogorov complexity is generally noncomputable, making $\Upsilon$, and thus AIXI, noncomputable with any finite compute budget. While there exist computable approximations of AIXI for a limited program length or finite compute budget, such solutions remains computationally infeasible beyond the most toy MDPs \citep{hutter2007universal, veness2011monte}. Further, the assumption of a universal prior over all MDPs serves as a rigid, subjective decision that limits the true generality of this definition. Ultimately, such \emph{top-down} definitions of general intelligence primarily offer insightful, theoretical sounding boards, but in practice, suggest no tractable path to what they describe.

In contrast, our \emph{bottom-up}, exploration-driven view of general intelligence traces back to earlier works on self-improving algorithms, which propose open-ended search mechanisms for finding guaranteed improvements to a problem-solving program. Gödel machines continually search the space of solver programs and attempt to prove whether each candidate is more general than the current solver, which is then replaced by the better candidate \citep{schmidhuber2007godel}. However, this method entails the exorbitant computational costs of brute-force proof search. Its successor, POWERPLAY instead seeks to find the simplest modifications to a program that results in a more general solver \citep{schmidhuber2013powerplay}, though this remains a costly search problem. Here, a simplicity constraint is enforced by limiting the compute budget available for searching for a new, yet unsolvable task and its corresponding modification to the solver. Similarly, prior work on Never-Ending Learning \citep{mitchell2018never} frames an open-ended learning process over an infinitely expanding knowledge graph, though its task space is limited to classification and inference tasks on this growing ontology. Our framework of open-ended exploration provides a  generalization of these approaches to data-driven agents operating over the full space of possible tasks. Importantly, active collection of new training data eases the costlier parts of search over the task space by explicitly considering the current capabilities and learning dynamics of the current model. By adapting the active collection criterion's relative weighing of learning potential, diversity, and grounding (as in Equation~\ref{eq:oel_exploration_criterion_grounded}), open-ended exploration mitigates the potential risk of the search process tunneling into limited portions of the task space, which may be completely divorced from the problems of interest to the system designer. Crucially, because Equation~\ref{eq:oel_exploration_criterion_grounded} rates novelty based on the agent's current predictive capacity and experiential history, exploration and the agent's capabilities interact in a mutually compounding manner: As the agent learns to achieve greater reward in more environments, maintaining exploratory novelty forces a gradual ratcheting of its mastery over an increasingly diverse set of environments. Conceivably, an agent that continues to explore and learn can eventually see its capabilities cover the set of all learnable skills within the domain of exploration.

A related vision for producing general intelligence stems from the open-ended evolution community within the field of artificial life (ALife), which is concerned with developing programs that replicate the emergent complexity characteristic of living systems \citep{boden1996philosophy, langton1997artificial}. Kickstarting a process that exhibits open-endedness---that is, the endless generation of novel complexity---is seen as a key requirement for achieving this goal. Such an open-ended process may then become an AI generating algorithm (AI-GA) \citep{clune2019ai}, by producing an ecosystem of increasingly complex problems and agents co-evolved to solve them. However, such a system may constitute a large population of agents specialized to specific challenges. Moreover, exactly how such a co-evolving system might be implemented in practice remains an open question. We propose open-ended learning as a path toward not only an open-ended process, but one resulting in a single, generalist agent capable of dominating (or matching) any other agent in relative general intelligence over time. In this way, open-ended learning bridges the search for open-ended emergent complexity in ALife with the quest for general intelligence in AI.

\section{Open-Ended Problems}
\label{sec:open_problems}

Open-ended learning requires SL datasets and RL environments that act as open-ended generators of both tasks and training experiences. To bring this discussion to a close and inspire future research, we describe a series of major open problems on the path to achieving such a system.

\subsection{Open-World Benchmarks}
The nascent subfield of Unsupervised Environment Design (UED) in the RL community recently begins to consider exploration over the entire space of environments \citep{paired, plr, robustplr}, and therefore over the full space of possible training data. Yet, ultimately, these preliminary works focus exclusively on static simulators that, despite being parameterized over a space of possible configurations, lack the ability to grow their potential configurations in open-ended ways, making them inappropriate for training an IGI. Similarly, popular benchmarks in SL, like GLUE \citep{glue, superglue} and ImageNet \citep{imagenet} typically center on a static dataset. The more recent StreamingQA benchmark explicitly focuses on the ability of a language model to adapt to a changing data distribution \citep{livska2022streamingqa}, but ultimately also relies on a static dataset. Other recent benchmarks propose problem settings more closely matching the signature features of open-ended learning---that of a continually expanding problem space and actively collected training data: BIG-bench invites researchers to collaboratively grow a repository of benchmark tasks for large language models \citep{srivastava2022beyond}. In a sense, BIG-bench can be seen as a slow, human-driven emulation of open-ended exploration over the task space. However, it does not truly capture the notion of open-ended exploration, as it only considers a finite set of human-designed tasks during training. Similarly, Dynabench seeks online collection of adversarial examples via humans and other models in the loop, allowing for open-ended expansion of the training data within the relevant task domain \citep{kiela2021dynabench}. However Dynabench is currently limited to active collection for a single, pre-specified human-designed task. A truly open-ended benchmark should test the ability of a learning process to continually discover new problems and acquire the necessary data to to solve them. Building such a benchmark requires finding solutions to several important, interrelated questions.

\emph{Q1. In which domain should we study open-ended learning?}

A primary challenge in creating an open-ended benchmark lies in defining the problem domain. ML has historically focused on individual problem domains that afford simple definitions, either through a hand-coded simulator or a representative, curated dataset. However, open-ended exploration requires a problem domain that both (1) continues to grow with the learner (i.e. offers learning potential and diversity) and (2) captures the quality of real-world tasks of interest (i.e. grounding). For this reason, current proposals for open-ended environments based on artificial open-world games such as MineCraft \citep{tessler2017deep,guss2019minerl,grbic2021evocraft,fan2022minedojo} and NetHack~\citep{kuttler2020nethack,samvelyan2021minihack}, while perhaps satisfactory for the ALife model of open-endedness based on unbounded novelty, are insufficient for the open-ended learning of a general intelligence. For example, a generally-capable MineCraft agent cannot be trusted to perform any task beyond the confines of the game, such as driving your car. This distinction is important: Neither open-endedness nor open-ended learning in itself implies a process that achieves general intelligence. Grounding to real world tasks of interest plays a fundamental role, as emphasized in Equation~\ref{eq:oel_exploration_criterion_grounded}, and current open-ended simulators do not capture a task space satisfying this constraint. One viable candidate is the Internet---itself, an open-ended system capturing the world's knowledge. Like our virtual assistant, an agent that freely explores the rich task space over the comprehensive ontology of content on the Internet will necessarily develop general intelligence capable of understanding our world and the actions, mediated by software, that must be taken to affect specific changes within it. Another possibility is to develop an embodied intelligence that can freely explore the real world, though this possibility will likely need to be supplemented by learning in simulation. Further, it remains an open question what the right subset of the problem domain---whether Internet or real world---provides the most effective starting tasks from which an open-ended learner can progress to increasingly general capabilities. For example, should the learner begin by seeking to understand rich dynamics about the world by exploring video-centric tasks or first develop a rich grounded understanding of language through conversational or information retrieval challenges?

\emph{Q2. How do we design scalable open-ended data generators?}

An equally important and complex challenge is that of designing a computational system capable of representing an endlessly growing set of tasks. While any task may be defined as a program implementing a decision process, naively storing and searching all such programs discovered in a non-parameteric fashion, as proposed in prior works \citep{schmidhuber2013powerplay}, is computationally infeasible. Open-ended learning requires a generative process capable of continually inventing new tasks, while storing only the most useful task designs in a compressed representation. Ideally, the number of parameters in such a generator grows much more slowly than the number of tasks represented. While we might imagine a large generative model, such as a code-generation model or world model, as playing the role of such a generator, it is unknown whether current model architectures and optimization methods are suitable for this kind of continual invention and compression.

\emph{Q3. How should an agent interface with an open-ended task space?}

In the long term, an open-ended learner must develop a general strategy for processing inputs from an increasingly diverse observation space and making decisions over an increasingly large action space. A generic interface between agent and environment requires adapting the input and output representations of the agent's underlying model to the task at hand. The adapted interface may take the form of tools invented by the agent, e.g. real or simulated hardware, or even a program. Such tools may even be passed on to other agents, which may further evolve the tool for new purposes. Recent work proposes that the invention of tools, which can mediate the interactions between other agents and the environment independently of the inventors, may play a pivotal role in the emergence of open-endedness \citep{lehman2022evolution}. The interface question is deeply related to the choice of domain for open-ended learning. In a virtual domain such as the web, the interface may be more abstract and programmatic in nature, while in a physically embodied domain, such as the real world, the interface may consist of sensors, dexterous actuators, and physical tool use. In each setting, we expect an open-ended learner to progressively innovate its own interface, reflecting the arc of human technology. 
\newpage
\emph{Q4. How do we measure the extent of open-ended learning?}

Even with the above problems addressed, there are no commonly-accepted measures for tracking the degree of open-ended learning achieved---that is, some measure of increasing capability. Previously proposed measures of open-endedness cannot be adapted for this purpose, as they focus on measuring novelty \citep{bedau1998classification, standish2003open, soros2014identifying}, rather than model capability. In general, such novelty and model capability are unrelated. For example, a process that evolves an agent across an endless range of mazes, while progressively growing the size of the agent's memory buffer, may score highly in novelty, but remains limited in capability. Performance-centric measures based on the number of tasks on which the agent experienced improved performance, such as the ANNECS metric \citep{wang2020enhanced}, suffer the same shortcoming: The learning process that fixates on the maze domain may see the agent struggle with new maze variations before solving them, thus propping up such measures without increasing general capabilities. One feasible option may be to simply track the diversity of tasks based on domain-specific criteria, but such a solution does not apply to all domains of interest. A general metric for open-ended learning would need to be domain-agnostic. Such a metric might consider both the agent's behavior in discovered tasks and task novelty based on a general task representation.

\vspace{-3mm}
\subsection{Active Collection}

Given a benchmark accommodating open-ended learning, there remain several major open problems centered on active collection---the means by which the learning process selects and gathers new training data. Existing UED methods perform a limited form of active collection by generating new simulator configurations, though limited to those represented within a fixed parameterization. Similarly, recent methods in SL perform active collection by actively generating new data based on data augmentations of existing data \citep{tran2019bayesian}. Open-ended exploration requires rethinking such approaches to search beyond a single, fixed parameterization of the data-generation process.

\emph{Q5. How do we determine what data to acquire next via active collection?}

Open-ended exploration requires searching for new training data across the entire data space. As argued in our construction of Equation~\ref{eq:oel_exploration_criterion_grounded}, this search process should be guided by a priori criteria, such as the informational worth, i.e. learning potential, of the collected data, and its similarity to specific tasks of interest. As the learning potential and diversity terms in the open-ended exploration criterion in Equation~\ref{eq:oel_exploration_criterion_grounded} are dependent on the current model, they present challenging non-stationary search objectives for active collection. Searching a latent representation of the data space may be required for efficient search, as well as the use of surrogate models to cheaply approximate the value of a datapoint under the search criterion. For example, this latent space might correspond to the input context to a Transformer-based generative model of the data or of data-generating programs. Existing methods based on latent-space optimization \citep{kusner2017grammar,lu2018structured,winter2019efficient,tripp2020sample} and quality-diversity search~\citep{gaier2018data,zhang2022deep,bhatt2022deep} make use of learned low-dimensional representations and surrogate models in this way, but assume stationary objectives. Moreover, it is currently unclear how similar criteria can be used to plan manual data collection procedures, such as hand-coded data-crawling scripts or test-time hand-offs to humans-in-the-loop. Further, we lack principled ways to predict the relative efficacy of training on actively collected data sourced manually, online, or offline. Lastly, it is unclear how the relative weighing between the terms in Equation~\ref{eq:oel_exploration_criterion_grounded} should evolve over time.

\emph{Q6. How can we scalably augment our training data?}

Efficient simulators enable RL agents to train on billions of data points in the span of days~\citep{petrenko2020sample, petrenko2021megaverse}. A promising direction to achieve similar computational advantages in SL, where we lack a simulator, is to directly learn a simulator of the training data. Such a simulator might consist of a generative model~\citep{jaderberg2014synthetic,bansal2019chauffeurnet,tripathi2019learning,such2020generative} or a partially-induced program that models the  data-generating process, e.g. a grammar modeled after sequential data~\citep{bouchard2016learning,kusner2017grammar,nye2020learning,chen2020compositional}. Such simulators can generate massive amounts of \emph{synthetic data}, serving as promising, parameterized regions of the data space over which to perform active collection, as well as for performing data augmentation of existing training data. 

\emph{Q7. How much prior knowledge should be used to ground exploration?}

Including a grounding term in the exploration criterion focuses active collection on parts of the data space resembling specific tasks of interest. Such a similarity constraint allows us to embed key inductive biases into the search process, e.g. if we want to make sure that any discovered simulator settings are consistent with specific physical laws. However, a strong grounding term can also force the search process to ignore important regions of the data space that may provide higher learning potential for the model, despite lacking a close resemblance to our exemplar data. In the evolutionary computing community, it is well known that novelty search can often lead to simpler, more effective solutions than objective-based optimization, through the discovery of useful stepping stones that would otherwise be overlooked \citep{lehman2011abandoning}.

\emph{Q8. How do we safely perform online collection?}

In prioritizing specific data points for training, active collection and prioritized training inherently distort the training distribution with respect to the ground-truth distribution in the real world. Such distribution shifts can result in biased models and thus harmful downstream effects when such models are deployed. Recent works characterize the bias that can result from prioritized training in both SL and RL, alongside methods for correcting for this bias \citep{farquhar2021statisticalbias, jiang2021grounding}. However, we lack principled means to understand how such biases may materialize in the open-ended learning setting, where data for tasks may be simultaneously, actively collected and thereby result in cross-task interference effects.
 
\subsection{Orthogonal Challenges}
We view these directions as the fundamental challenges of exploration. Other open problems, focused on optimization, directly impact an agent's capacity to explore, but are not core to the generalized exploration problems we consider. These orthogonal challenges, including catastrophic forgetting \citep{robins1995catastrophic, french1999catastrophic, kirkpatrick2017overcoming} and the challenges of gradient-based optimization of nonconvex objectives \citep{pascanu2013difficulty,schulman2015gradient,bottou2018optimization} and differentiable games \citep{balduzzi2018mechanics, mazumdar2019finding, mazumdar2019policy}, have historically been major areas of focus for the ML community. An agent's capacity to explore hinges on its ability to recognize novel data, which assumes a mastery of past experiences. More efficient optimization will thus generally benefit exploration. Similarly, finding new challenges entails retaining solutions to those already mastered. Exploration thus stands to directly benefit from methods addressing catastrophic forgetting.

\section{Conclusions}
\label{sec:conclusion}

The ML research community has made incredible advances over the past decade. However, outside of limited pockets of research, these advances have been developed primarily under the assumption of a static dataset, or otherwise, algorithmically generated data of limited description length, largely beyond the control of the learning agent. Thus, the overarching perspective of the ML community has largely ignored the generalized problem of exploration in learning: The learning agent must perform active collection, that is, it must both determine and acquire the data most informative for expanding its present capabilities on a continual basis, by considering three key aspects of the data: learning potential, diversity, and grounding to real tasks of value. We have argued that formulating this notion for existing ML methods requires rethinking the prevailing paradigm of exploration developed in RL, expanding its scope from a single, static environment simulator to the full data space containing information relevant to an unbounded set of possible tasks. Such an open-ended exploration process serves as the data-centric driver of open-ended learning. We believe open-ended exploration provides a more viable, bottom-up path toward general intelligence than alternative top-down approaches proposed in the past. On a short time horizon, independent of AGI, most ML systems deployed in an open world setting, with real users and peers, must perform some notion of open-ended exploration and learning, or risk fading into irrelevance. On a longer time horizon, the arc of such real-world ML systems is toward increasingly general capabilities, as continued improvement for most open-world tasks, from question-answering to driving, requires ever greater levels of sophistication in reasoning. Of course, our proposed, data-driven path to general intelligence is not without deep, unresolved open questions which we also discussed at length. 

We emphasize that the goal of this discussion is not to determine an exact solution to the problem of open-ended exploration. Rather, we aim to sketch the outlines of the problem and its subcomponents, so that the community can begin to fill in the missing pieces under a common conceptualization. As our learning algorithms become ever more proficient at modeling any dataset, we stand to gain the most from thinking more deeply about what data we feed these algorithms. Open-ended exploration, in its essence, is the principled and continual collection of training data to maximize a model's performance across the data space. By directing the research community's own exploration to this much uncharted frontier, we are confident in making progress toward ever more general AI systems.

\acksection{
We thank Michael Dennis, Gary Marcus, Shane Legg, Yoshua Bengio, Jürgen Schmidhuber, Robert Kirk, Laura Ruis, Akbir Khan, and Jakob Foerster for useful feedback on earlier drafts of this work.
}

\bibliographystyle{abbrv}
\bibliography{refs}

\end{document}